\documentclass[conference]{IEEEtran}
\IEEEoverridecommandlockouts
\usepackage{times}  
\usepackage{helvet} 
\usepackage{courier}  
\usepackage[hyphens]{url}  
\usepackage{graphicx}
\usepackage{booktabs}
\usepackage{mathtools}
\usepackage{amsmath}
\usepackage{algorithm,algorithmicx,algpseudocode}
\usepackage{multirow}
\usepackage[export]{adjustbox}
\usepackage[table]{xcolor}
\usepackage{array}
 \usepackage{setspace}
 \usepackage{cite}

\newcolumntype{C}[1]{>{\centering\let\newline\\\arraybackslash\hspace{0pt}}m{#1}}

\newcommand{\etal}{\textit{et al. }}

\title{When Explainability Meets Adversarial Learning: \\ 
Detecting Adversarial Examples using SHAP Signatures}

\author{Gil Fidel\qquad Ron Bitton \qquad Asaf Shabtai \\ 
Department of Software and Information Systems Engineering \\ 
Ben-Gurion University of the Negev}

\begin{document}

\maketitle

\begin{abstract}
State-of-the-art deep neural networks (DNNs) are highly effective in solving many complex real-world problems. 
However, these models are vulnerable to adversarial perturbation attacks, and despite the plethora of research in this domain, to this day, adversaries still have the upper hand in the cat and mouse game of adversarial example generation methods vs. detection and prevention methods.
In this research, we present a novel detection method that uses Shapley Additive Explanations (SHAP) values computed for the internal layers of a DNN classifier to discriminate between normal and adversarial inputs.
We evaluate our method by building an extensive dataset of adversarial examples over the popular CIFAR-10 and MNIST datasets, and training a neural network-based detector to distinguish between normal and adversarial inputs.
We evaluate our detector against adversarial examples generated by diverse state-of-the-art attacks and demonstrate its high detection accuracy and strong generalization ability to adversarial inputs generated with different attack methods.
\end{abstract}


\section{Introduction}
In recent years, deep neural network (DNN) learning algorithms have been widely used to solve a variety of complex problems.
Their greatest impact has been seen in fields such as image classification, object recognition, natural language processing, and malware detection.

Despite their outstanding performance - often outperforming human experts - DNNs have been shown to be vulnerable to adversarial perturbations.
First discovered by Szegedy~\etal~\cite{szegedy2013intriguing}, adversarial perturbations are slight modifications of DNN input that cause misclassification.
For example, in the domain of image classification - such modifications could be small adjustments in pixel colors that are imperceptible to humans, yet cause state-of-the-art classifiers to produce output arbitrarily chosen by an attacker.

Since then, extensive research has been conducted on adversarial examples focusing on four major directions: \textit{adversarial example generation methods}~\cite{moosavi2016deepfool,athalye2017synthesizing,carlini2017towards,kurakin2016adversarial,madry2017towards}, \textit{defenses for increasing the robustness of DNN models against adversarial examples}~\cite{papernot2016distillation,dziugaite2016study,shaham2018understanding}, \textit{adversarial example detection}~\cite{metzen2017detecting,feinman2017detecting,grosse2017statistical,xu2017feature,katzir2018detecting,pang2018towards,roth2019odds}, and \textit{understanding the nature and root causes of adversarial examples}~\cite{shamir2019simple,ilyas2019adversarial,goodfellow2014explaining,madry2017towards}.

Currently, attackers are still ahead in their arms race with the defenders, with state-of-the-art defenses falling short in the face of advanced adaptive attacks~\cite{carlini2017adversarial}.
Thus, the ability to effectively detect adversarial examples remains an open problem.

Another, seemingly unrelated, yet notable shortcoming of DNN models, is the difficulty in explaining the rationale, or even providing supporting evidence to justify their decisions. 
This poses a significant obstacle to their adoption in production-grade contexts~\cite{gunning2017explainable}.
For this reason, extensive research efforts are being invested in the field of explainable artificial intelligence (XAI) to improve the ability of humans to interpret the decisions made by DNN and other machine learning models~\cite{adadi2018peeking,samek2017explainable,zhang2018visual,shap2017}.

We hypothesize that a deep connection exists between model explainability and adversarial examples.
Intuitively, a well explained model should be fairly robust to adversarial perturbations, since adversarial input would result in the emergence of anomalous explanations for the model's decision.
Our goal in this paper - is to uncover and utilize this connection to advance the state of the art in adversarial example detection.
We present and evaluate a novel adversarial example detection method that applies the SHAP explainability technique~\cite{shap2017} on the penultimate layer of a DNN to create ``XAI signatures" which are fed into our detector.

We evaluated our proposed method using the CIFAR-10~\cite{krizhevsky2009learning} and MNIST~\cite{lecun1998gradient} datasets, generally following the strict adversarial defense evaluation guidelines set forth by Carlini~\etal~\cite{carlini2019evaluating}.
The evaluation results show that our method is highly effective in detecting adversarial examples (AUC \textasciitilde 97\%) and generalizes well across different adversarial example generation algorithms. 
The excellent generalization results support our hypothesis that our method captures an intrinsic property of adversarial examples.
In contrast to prior detectors, we evaluate ours against a wider range of adversarial attacks (both white-box and black-box), including the strongest known attacks, and achieve very high detection ROC-AUC scores in both scenarios: adversarial examples generated by attack methods the detector was trained on, and adversarial examples generated by attack methods that the detector was not trained on.

To summarize, our main contributions in this study are two-fold:
(1) we introduce a novel adversarial example detection method with an impressive detection performance and demonstrate its high effectiveness against a diverse range of adversarial attacks; and (2) we make a first step towards uncovering a deep link between adversarial learning and explainable AI.

\section{\label{sec:background}Background}

\subsection{Adversarial Attacks}
Attacks against machine learning classifiers, denoted as adversarial machine learning, occur in two main phases of the machine learning process: during model training, also known as poisoning, and during the classification phase, also known as evasion attack. 
A poisoning attack can be performed by inserting crafted malicious samples to the training set as part of the baseline training phase of a classifier.
In this research we focus on evasion attacks, and specifically, detecting adversarial examples.
An evasion attack involves modifying the analyzed sample's features to evade detection by the model. 
Such samples are called adversarial examples \cite{szegedy2013intriguing}.

Given a classifier $f: R^n \rightarrow C$ mapping a floating point vector of an input sample to a class in the set of possible target classes, an input sample $x \in R^n$, and a correct class label $c$, we call $\delta \in R^n$ an \textit{adversarial perturbation} and $x' = x + \delta$ an \textit{adversarial example} if:

\begin{equation}
    \begin{aligned}
        f(x') \neq c, \\
        s.t: ||\delta|| < \epsilon        
    \end{aligned}
\end{equation}
where $||\cdot||$ is a distance metric and $\epsilon > 0$ is the maximum allowed perturbation size which is set to a small positive value to constrain the perturbation s.t. that resulting adversarial example is indistinguishable from the original sample to the naked eye. 
Although the perceived difference between the original and perturbed samples is difficult to estimate, the distance metrics used in most adversarial attacks are $L_0$ (the number of input features changed), $L_2$ (standard euclidean distance) and $L_\infty$ (maximum difference of any single feature), each one providing a good, albeit different, approximation of the perceived difference. 

The algorithm or method used to generate the adversarial example is often referred to as an \textit{adversarial attack}. 
A \textit{targeted} attack generates an adversarial example that gets classified as a specific, attacker defined, target class, whereas a non targeted attack merely causes a misclassification to \textit{any} incorrect target class.
Many adversarial example generation methods have been invented in recent years. 
Some of the most notable, which are among the ones we use for our evaluation include:

\begin{itemize}
    \item \textit{Fast Gradient Sign Method (FGSM)} \cite{goodfellow2014explaining} - a basic technique that involves taking a single step in the input space in the direction of the gradient of the model's cost function with magnitude equal to the max allowed perturbation norm ($\epsilon$). 
    \item \textit{Basic Iterative Method (BIM)} and its variation \textit{Projected Gradient Descent (PGD)} \cite{kurakin2016adversarial,madry2017towards} - a natural extension of FGSM that takes multiple FGSM-like steps with smaller step sizes adding up to less than the maximum allowed perturbation size ($\epsilon$). 
    \item \textit{Carlini \& Wagner (C\&W)} \cite{carlini2017towards} - an attack that formulates the problem of finding an adversarial examples as an optimization problem with a cleverly chosen loss function tailored for each metric.
\end{itemize}

\noindent Adversarial attacks can be further divided to \textit{white box} and \textit{black box}.
In the \textit{white box} scenario, an attacker has full access to the attacked classification model, including its internal structure and parameters/weights. 
In \textit{black box} attacks, the attacker can only feed the model with inputs and observe the outputs but doesn't have access to its internal state. 
An important property of many adversarial attacks is \textit{transferability}~\cite{papernot2017practical}.
According to this property adversarial examples generated against one model can fool other models as well.
This allows converting a white-box into a black-box attack by training a surrogate model and generating an adversarial example against it that can be successfully fool the original model.

\subsection {Adversarial Defenses}
Previously proposed defense methods against adversarial attacks can be categorized as methods that aim at improving the robustness of the trained model to adversarial attacks and methods that aim at detecting adversarial examples.
Methods for generating robust models include adversarial training~\cite{goodfellow2014explaining}, Defensive Distillation~\cite{papernot2016distillation}, Gradient Obfuscation, Feature Squeezing \cite{xu2017feature} and more.

In this research we propose a method for detecting adversarial examples.
Most previously proposed methods for detecting adversarial examples attempted to identify irregularities in the input data, or in the internal behavior (e.g., internal layer activations) of the model, while others take a more active approach that involves transforming the inputs~\cite{xu2017feature} or modifying the training process~\cite{roth2019odds} to improve the detector performance. 

\subsection {Understanding adversarial examples}
Ilyas \etal~\cite{ilyas2019adversarial} argue that the existence of adversarial examples is actually an intrinsic property of the dataset itself.
They introduce that notion of \textit{robust} and \textit{non-robust} features. 
Non-robust features are highly predictive, yet very fragile and prone to change drastically due to small perturbations of the input. 
Robust features, on the other hand, are features that are both highly predictive and do not change easily by small change in the input. 
One can think of robust features as features that capture some important, high-level, feature of the target class - such as the presence of wheels and windows for cars, whereas non-robust features are seemingly random patterns that aren't noticeable by human beings, but emerge as highly predictive during the training process. 
Ilyas \etal show that the existence of adversarial examples, as well as their transferability across different classification models, naturally arises from the existence of non-robust features since they allow small perturbations in the input to cause major changes in value of these highly-predictive features.

\subsection {Explainable AI}
Explainable AI (XAI) is an emerging researched field in machine learning with the purpose of allowing users to understand, trust, and effectively manage the next generation of AI solutions~\cite{gunning2017explainable}.
Most of the XAI methods developed in recent years are meant to explain supervised machine learning models. 
For example, the LIME~\cite{ribeiro2016should} method introduced for explaining the prediction using a local model; the DeepLIFT method~\cite{shrikumar2017learning}, which uses back propagation through all of the neurons in the network to explain the output; and the SHAP~\cite{shap2017} method, which is a unified approach that aims to explain the model output using shapely values - a concept borrowed from game theory where it is used to calculate the relative contributions of different players in a coalition. In the context of XAI, they are used for estimating the contribution of a specific input or neuron to a model's decision.
The need to explain the output is especially important in anomaly detection based on deep learning models, because usually in this case not all of the anomaly types are known (labeled).
In this research we use the SHAP DeepExplainer method, which is a variation of the SHAP algorithm that is specifically optimized for explaining DNN models.

\section{\label{sec:related_work}Related Work}
Previous works suggested methods for detecting adversarial examples.
A summary of these works is presented in Table~\ref{table:relatedworks}.
The table presents a succinct summary of the detection concept and evaluation setup and results of each detector. 
In addition, we summarize the most important pros and cons of each detector.
In this research we propose a novel approach for detecting adversarial sample which was not proposed before.
In addition, we conduct a more comprehensive evaluation by checking the models ability for cross attack generalization as well as evaluating on a larger quantity of diverse attack types, thus providing a higher confidence in our model's ability to adapt to real-world challenges.

\begin{table*}[h!]
\scriptsize
\centering

\begin{tabular}{|C{0.05\linewidth}|C{0.15\linewidth}|C{0.08\linewidth}|C{0.08\linewidth}|C{0.16\linewidth}|C{0.16\linewidth}|C{0.15\linewidth}|}

\hline
\textbf{Ref} & \textbf{Concept} & \textbf{Datasets} & \textbf{Attacks} & \textbf{Main Results} & \textbf{Pros} & \textbf{Cons} \\
\hline
\noalign{\hrule height 1pt}

\cite{metzen2017detecting} & Binary detector fed by activation of internal layers & CIFAR-10, MNIST, IMAGENET-10 & FGSM, BIM ($L_2$, $L_\infty$), DeepFool ($L_2$, $L_\infty$) & Accuracy: between 0.79 and 0.97 (average 0.87) when training and testing on the same attack; cross attack scores are much lower & Evaluating transferability between different attacks and perturbation budgets & Relatively low detector accuracy, especially in generalization scenario; evaluating on a small set of attacks \\
\hline

\rowcolor[gray]{0.9}
\cite{katzir2018detecting} & Cluster activations of internal layers and classify adversarial examples based on movements between different clusters & CIFAR-10, MNIST & C\&W ($L_2$)  & CIFAR10 – AUC=0.92 (0.95 for correctly classified only) ; MNIST – AUC=0.91 & Good performance against the strong C\&W attack & Evaluating against the C\&W $L_2$ attack only (might not perform well against other attacks) \\

\hline

\cite{feinman2017detecting} & Extract features to construct two unsupervised detectors and also an ensemble of them: 1. Density estimated of feature space of last hidden layer 2. Bayesian uncertainty estimates, available in dropout neural networks.  
& MNIST, CIFAR-10, SVHN & FGSM, BIM, JSMA, C\&W & CIFAR-10: Average AUC of 0.8554 & Unsupervised approach (no need for adv samples for training)
& Low detector performance (AUC), evaluating on a weak CIFAR-10 classifier \\
\hline

\rowcolor[gray]{0.9}
\cite{grosse2017statistical} & Statistical tests on raw inputs & MNIST, DREBIN 
, MicroRNA
& FGSM, JSMA (on MNIST) & Detection rate: FGSM 99\% JSMA 80\%  & Generic approach (applicable to different domains and model types such as SVM and Decision Trees)
& Evaluating on weak attacks
; low performance on adv samples generated using JSMA \\
\hline

\cite{pang2018towards} & Density estimate anomaly detector like~\cite{feinman2017detecting};
additionally, change training loss function of defended model from cross entropy to Reverse Cross Entropy (RCE)
to improve detection rates & MNIST, CIFAR-10 & FGSM, BIM, ILCM, JSMA, C\&W & AUC: FGSM 99.7, BIM 100.0, ILCM 84.2, JSMA: 85.8, C\&W: 95.3; RCE model: C\&W 91.8, ILCM 93.9, JSMA 95.4, C\&W 98.2, FGSM 99.7, BIM 100 & Good performance on RCE trained model; unsupervised approach (no need for adv samples for training) & Requires customizing the training of the defended model \\
\hline

\rowcolor[gray]{0.9}
\cite{xu2017feature} & Decrease input resolution (bit depth reduction, local smoothing), making adv attacks more difficult and detection easier by comparing model output for original and transformed inputs
& CIFAR-10, MNIST, IMAGENET & FGSM, BIM, C\&W ($L_0$, $L_2$, $L_\infty$), DeepFool, JSMA & CIFAR-10: TPR of 0.845 (FPR=0.05) & Evaluating on a diverse set of adversarial attacks; good detection performance against the strong C\&W attack & Weak performance against the BIM, DeepFool, JSMA, FGSMS attacks (relatively weak overall detection performance) \\
\hline

\cite{roth2019odds} & Based on statistical differences in the distribution of logits of f(x+$\epsilon$) 
& CIFAR-10, IMAGENET & Train on PGD $L_\infty$ and Eval on PGD $L_\infty$, PGD $L_2$, C\&W $L_2$  & PGD $L_\infty$: 99.1\% TPR / 0.2\% FPR, PGD\L 2: 96.1\% / 1\% C\&W L 2: 91.6\% / 4.8\% & High detection performance; generalization across adversarial attacks & Not evaluating on a diverse set of attacks \\
\hline

\noalign{\hrule height 1pt}

\end{tabular}
\caption{Summary of related works.}
\label{table:relatedworks}
\end{table*}

\section{Robustness Through Explainability}
\begin{figure*}
\centering
  \includegraphics[scale=0.95]{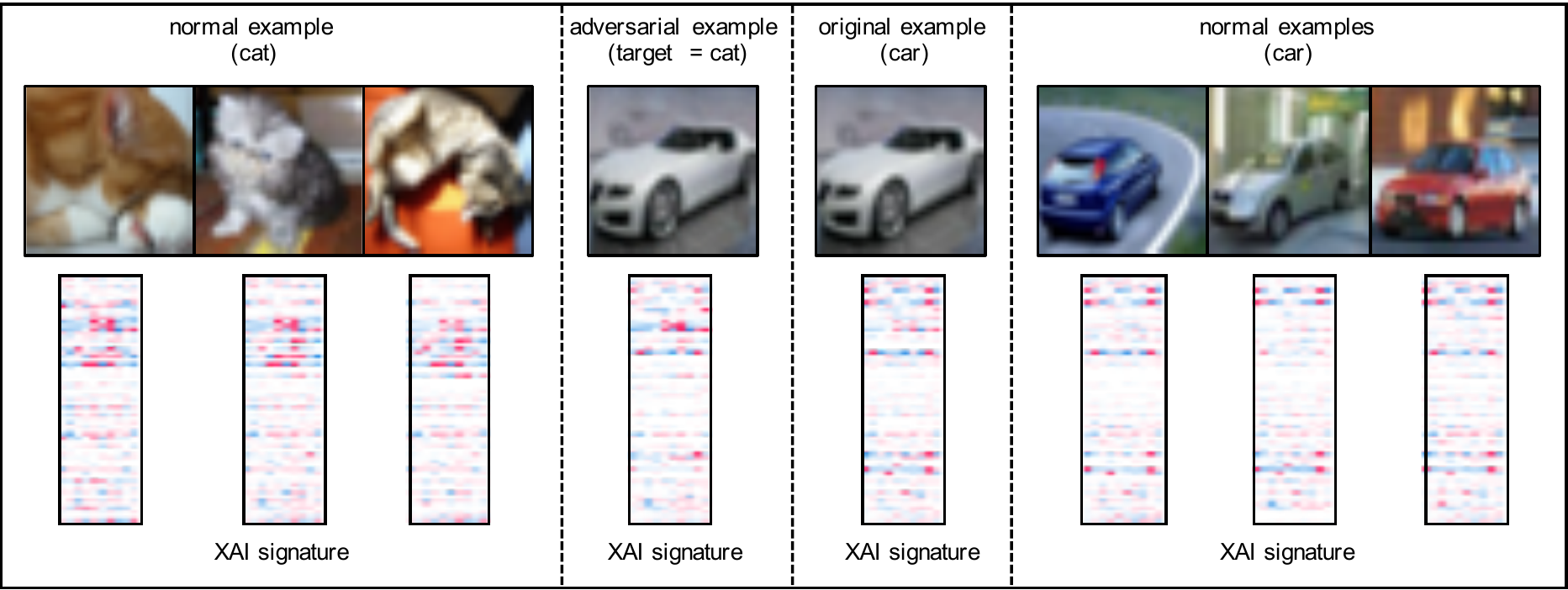}
  \caption{Illustrating the XAI signatures of examples from different classes, as well as original and adversarial example.}
  \label{fig:feature_exploration}
\end{figure*}

Adversarial evasion attacks change the values of non-robust features, while largely leaving robust features intact~\cite{ilyas2019adversarial}.
This is because applying an effective modification to robust features requires significant changes to the input.
Consequently, we hypothesize that we should see different patterns in the importance of robust vs. non-robust features in the classification of normal and adversarial inputs, with the latter relying more heavily on non-robust features.
We try to leverage this hypothesized property of adversarial examples by utilizing explainable AI methods (XAI) for interpreting model predictions.

Consequently, for each input to be classified as adversarial or normal, we utilize SHAP~\cite{shap2017} to compute the importance scores of the neurons of the penultimate layer of the classification model.
Then, we use these importance scores as features for our adversarial example detector.
The reason for interpreting the penultimate layer (instead of the input layer for instance) is because the neurons of this layer actually form high-level features of the original classification model~\cite{goodfellow2016dlbook}.

Figure~\ref{fig:feature_exploration} provides an illustration that supports our hypothesis.
On the left side of the figure, we can see three normal examples of the same class ``cat", and on the right side we can see three normal examples of another class ``automobile".
In the middle of the figure we can see a normal (original) example from the class ``automobile" and a perturbation of that example after applying a targeted (target class ``cat") PGD $L_2$ attack~\cite{madry2017towards}.
Below each image (example) we present the SHAP XAI signature of the image, such that each pixel in the signature at coordinates ($row$ = $i$,$col$ = $j$) contains the SHAP value of neuron $i$ for target class $j$. 
Red pixels denote positive contributions of their respective neurons for steering the model's decision towards the respective target class, whereas blue pixels denote a negative contribution, steering the model away. 
The intensity of the color denotes the magnitude of the positive or negative contribution, with white/transparent pixels denoting no contribution at all.
From a birds eye view of the figure it can be observed that the XAI signatures of images of the \textit{same} class (automobiles or cats) are similar, while different classes have different XAI signatures.
In can be also observed that although the original and perturbed automobile examples look the same, their XAI signatures are different. 
A closer look, however, uncovers even more intriguing properties: The normal car contains five relatively distinctive rows in their XAI signatures (two near the top, two near the bottom and one closer to the middle). 
Moreover, the bright red pixels in these rows are located in columns 1 and 9 which correspond to the target classes ``automobile" and "truck". 
On the left hand side of the figure, the three normal cat examples share a similarly looking lump of red pixels in the upper middle part of the XAI signature. 
Moving over to the adversarial automobile example, we can see that it shares two of the five distinctive rows with the normal cars, and a red lump in the upper middle with the cats. 
Thus, a mixture of ``automobile" and ``cat" features plays an important role in the decision of the underlying classifier for this adversarial example.
Although this is merely a speculation and further research is required to draw strong conclusions, but we hypothesize that this behaviour is perfectly aligned with the notion of robust and non-robust features: The two distinctive rows that the adversarial attack failed to alter correspond to robust features of the ``automobile" class, whereas the remaining three rows that did disappear correspond to non-robust ``automobile" features. 
Likewise - the part of the red lump that transferred from the normal cats to the adversarial one correspond to non-robust ``cat" features, while the part of the lump the didn't transfer corresponds to robust ``cat" features.

To explore the dataset in more depth, we trained a UMAP~\cite{mcinnes2018arXivUMAP} dimensionality reducer on the train set and used it to project the test set onto the embedding space of the train set.

\begin{figure}[h]
  \centering
  \includegraphics[scale=0.44]{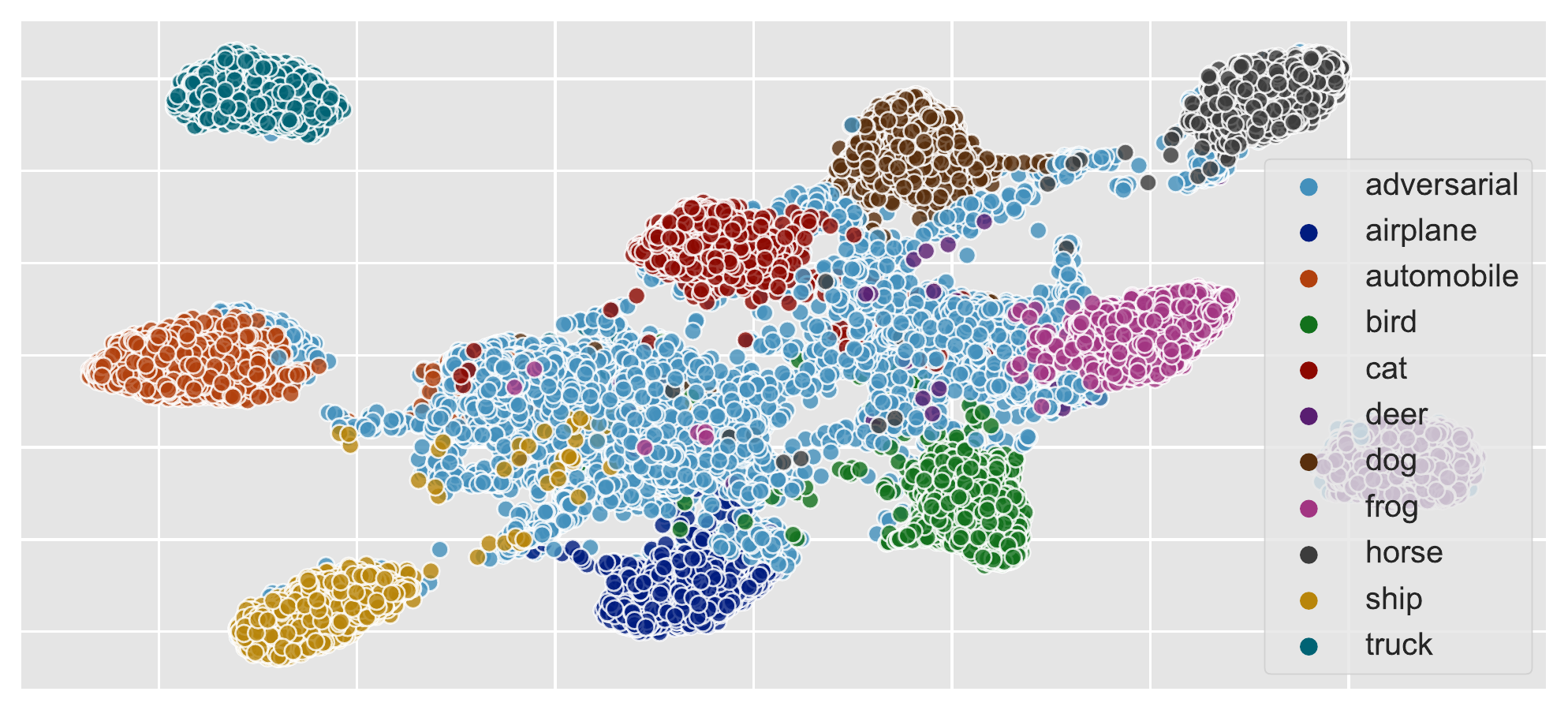}
  \caption{UMAP visualization of XAI signatures for normal and adversarial examples (CIFAR-10)}
  \label{fig:RQ1-CIFAR10-UMAP}
\end{figure}

\begin{figure}[h]
  \centering
  \includegraphics[scale=0.44]{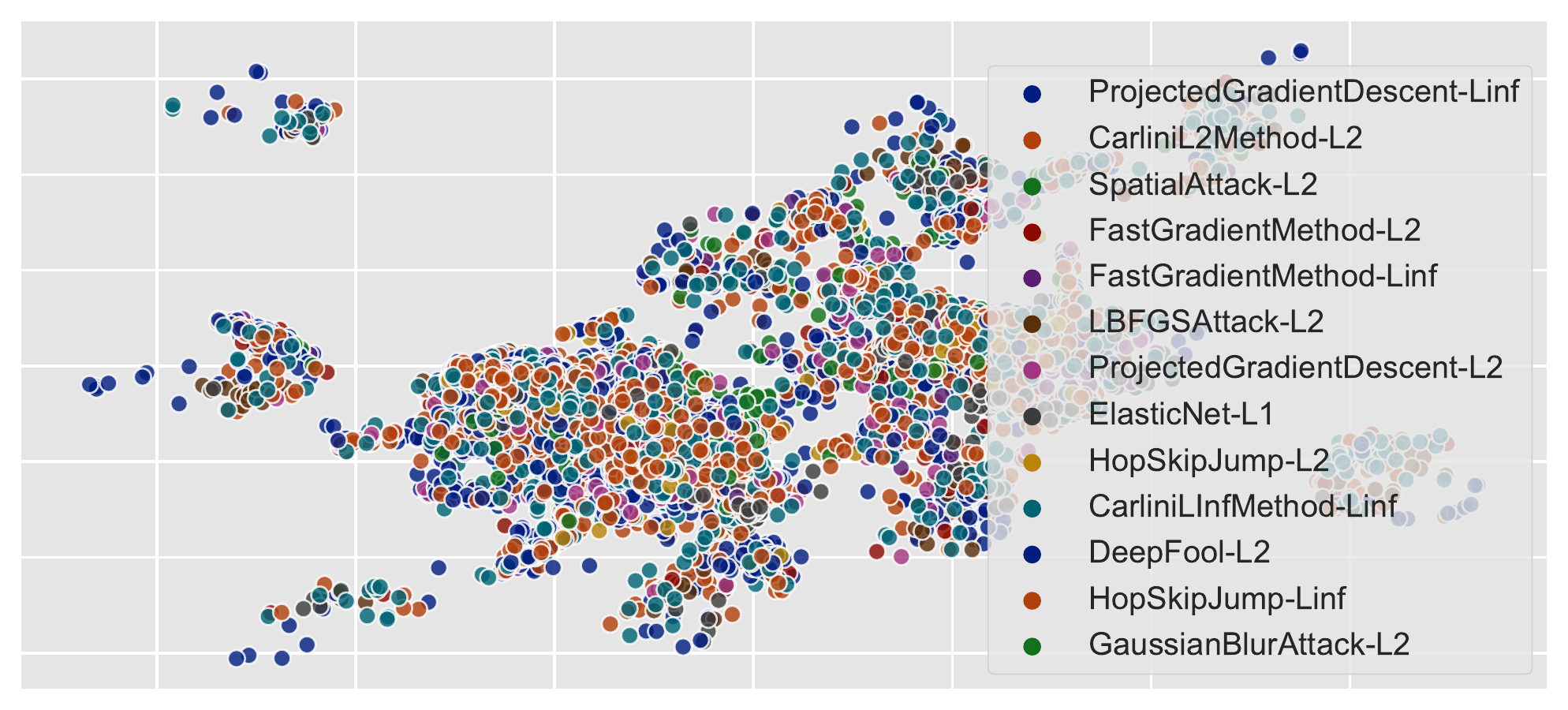}
  \caption{UMAP visualization of XAI signatures for different adversarial examples (CIFAR-10)}
  \label{fig:CIFAR10-UMAP-ADV-ONLY}
\end{figure}

In Figure \ref{fig:RQ1-CIFAR10-UMAP} we can clearly see ten distinct clusters of normal samples, one for each target class, and one cluster containing adversarial examples. From this separation we deduced that the computed SHAP values would make good features for our detector.
Figure \ref{fig:CIFAR10-UMAP-ADV-ONLY} shows only the adversarial examples, projected onto the same embedding space as in Figure \ref{fig:RQ1-CIFAR10-UMAP}. 
The wide spatial dispersion of different types of adversarial examples all around the clusters of adversarial examples hints that we could expect our detector to generalize well when tested on types of adversarial examples it wasn't trained on.

\section{Proposed Method \label{sec:method}}
\begin{figure*}
\centering
  \includegraphics[scale=0.36]{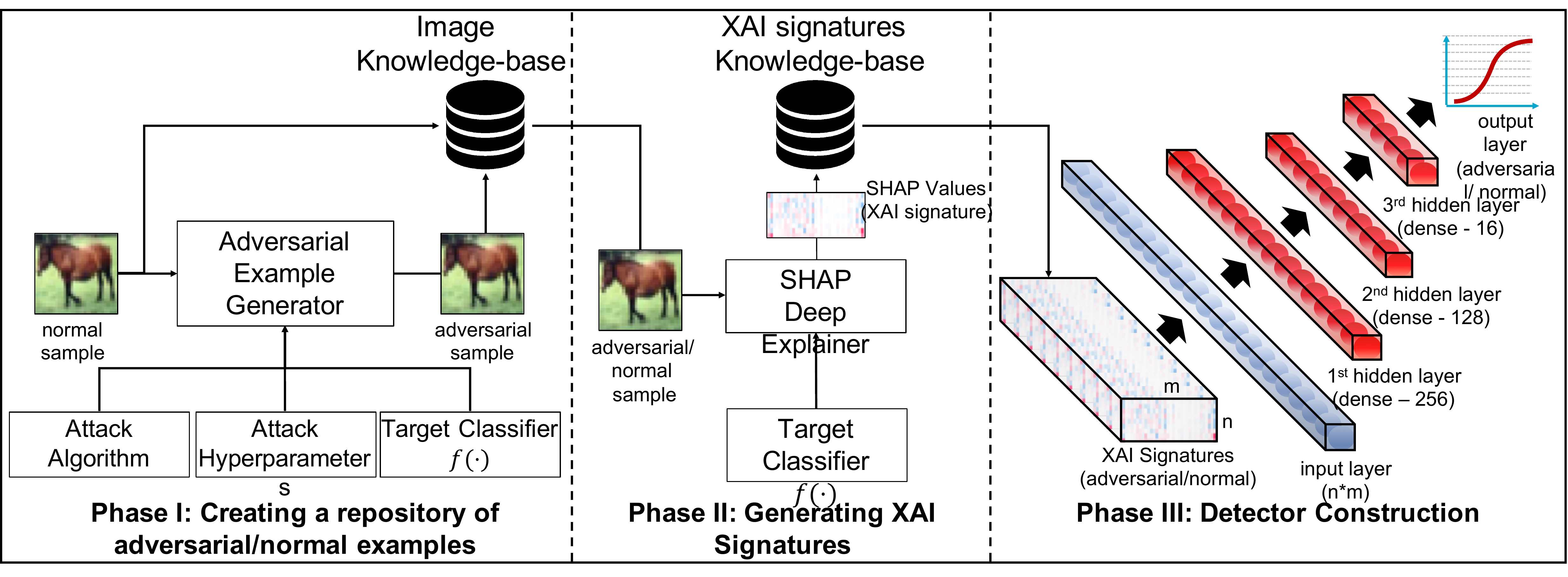}
  \caption{Detector training process.}
  \label{fig:process}
\end{figure*}

The proposed solution consists of three main phases (Figure~\ref{fig:process}): creating a repository of normal and adversarial examples, generating XAI signatures, and detector construction.

\subsection{Notation}
\begin{itemize}
    \item $f(\cdot)$ - a neural network based classifier
    \item $f^{[i]}(\cdot)$ - the output of the $i^{th}$ neural network layer ($0 \leq i \leq l$), where $f^{[0]}(\cdot)$ is the input layer and $f^{[l]}(\cdot)$ is the final softmax output.
    \item $x$ - input vector
    \item $Y(x)$ - ground truth label of $x$
\end{itemize}

\subsection{Creating a repository of normal and adversarial examples}
In this phase, a repository of normal and adversarial samples is generated.
The normal examples are randomly sampled from the dataset used to train $f(\cdot)$.
The adversarial examples are generated by applying a variety of state-of-the-art adversarial attack algorithms on $f(\cdot)$.
When generating adversarial examples, it is crucial to investigate the attack's hyperparameters such as the distance metrics (e.g., $L_0$, $L_1$, $L_2$, $L_{\infty}$), perturbation budget, number of iteration, attack step and etc)~\cite{carlini2019evaluating}.
Fuzzing over the various hyperparameters produces different types of perturbations (i.e., attacks) and consequently increases the generalization capability of the detection model.

Algorithm~\ref{alg:attack_gen} outlines the process of generating a representative set of adversarial examples.
As can be seen, in each iteration of the algorithm we randomly select: a normal sample from the dataset used to train the classification model (line 12); a combinations of attack method, distance metrics and attack preferences (lines 13-15); and a target class, which is different from the ground truth (line 16). 
Then, for each tuple (consisting of: \textit{ the sample, distance metric, attack preferences, target class and classification model}), we execute the attack to generate an adversarial example (line 17). 
If the attack ends successfully (i.e, the classification of the perturbation using the targeted classifier equal to the target class) we store it in our repository.

\begin{algorithm}[t]
\caption{Generating Adversarial Examples}\label{alg:attack_gen}
    \begin{algorithmic}[1]
        \State\textbf{Inputs:}
            \State $\qquad$ $X_{normal} \gets$ sampled normal examples
            \State $\qquad$ $L \gets$ set of possible labels
            \State $\qquad$ $M \gets$ set of attack methods
            \State $\qquad$ $D \gets$ set of distance metrics
            \State $\qquad$ $P(m) \gets$ set of preferences for attack method $m\in M$
            \State $\qquad$ $f(\cdot) \gets$ target classifier
            \State $\qquad$ $i \gets$ number of samples to generate
        
        \Procedure{GenerateAdversarialExamples($X_{normal}$ , $L$, $M$, $D$ , $P(m)$ ,$f(\cdot)$, $i$)}{}
            \State $\qquad X_{adversarial}  \gets \phi$
            \While {$i>0$}
                \State $x \gets RandomSample(X_{normal})$
                \State $m \gets RandomSample(M)$
                \State $d \gets RandomSample(D)$
                \State $p \gets RandomSample(P(m))$
                \State $target \gets RandomSample(L \cap Y(x))$
                \State $x^{*} \gets m(x,d,p,target,f(\cdot))$
                \If {$f(x^{*}) == target$}
                    \State $X_{adversarial} \gets X_{adversarial} \cup x^{*} $
                \EndIf
                \State $i \gets i-1$
            \EndWhile
        \State \Return $X_{adversarial}$
        \EndProcedure
    \end{algorithmic}
\end{algorithm}

\subsection{Generating XAI signatures}
In this phase, we utilize SHAP to generate an XAI signature for each sample in the dataset (both normal and adversarial).
Specifically, we apply the SHAP DeepExplainer~\cite{shap2017} to interpret the neurons of the penultimate layer $f^{[l-1]}(\cdot)$.
The outcome of this application is $n$ SHAP values for each output in $f^{[l-1]}(\cdot)$, where $n$ represent the number of target classes (i.e., SHAP produces a single value for each output and class).
The XAI signature of a given sample is defined as the concatenation of all SHAP values into a flat floating-point vector (i.e., the size of each signature is  to $n * |f^{[l-1]}(\cdot)|$). 
Normal signatures are used as a baseline for modeling the behavior of the normal decision-making procedure within the activation space.
Attack signatures are used for modeling the effect of different perturbations on the decision-making procedure.

It should be mentioned that in a production implementation of this approach, the repository should be updated continuously with attack signatures generated using newly discovered attacks.
Maintaining an updated repository will improve the performance of the proposed method in detecting new attack classes.
Nevertheless, a good detector must be able to generalize known attacks in order to detect unknown attacks (we discuss this topic in Section~\ref{sec:eval}).

\subsection{Training the detector}
In this phase, we train a supervised binary detector to discriminate between normal and adversarial samples, based on their XAI signatures.
We use the SHAP values from our generated dataset as the samples' features to train the classifier.
Any standard supervised model can be used to train the detector based on these features.
In this research we trained a fully connected feed forward neural network detector.
At inference time, given a sample to classify as normal or adversarial, we compute the sample's SHAP values (in the XAI signature phase) and feed the output into our binary classifier to classify the sample as adversarial or normal.

\section{Evaluation \label{sec:eval}}
In our evaluation we aimed to answer the following two research questions: 

\noindent\textbf{RQ1}: What is the baseline performance of the detector when the train and test sets contain adversarial examples generated using the same attacks and parameters?

\noindent\textbf{RQ2}: Can the detector generalize to adversarial examples generated by attacks that were unknown during training?
    
\subsection{Evaluation setup}
We evaluated our proposed detection method using the following two image classification use cases:

\noindent \textbf{[CIFAR,ResNet56]} The \textit{CIFAR-10} dataset~\cite{krizhevsky2009learning} with the \textit{ResNet-56} classification model~\cite{he2016deep}. 
    The model achieves a 93.39\% accuracy on the CIFAR-10 test set.
    
\noindent \textbf{[MNIST,CNN]} The \textit{MNIST} handwritten digits dataset~\cite{lecun1998gradient} with a model architecture taekn from the Keras MNIST example (https://keras.io/examples/mnist\_cnn/). 
    The model achieves a 99.25\% accuracy on the MNIST test set.

\subsubsection{Adversarial example generation.}
We generated adversarial examples using both the Foolbox~\cite{rauber2017foolbox} and Adversarial Robustness Toolbox~\cite{art2018} frameworks.
The SHAP explanations were generated using the SHAP framework~\cite{shap2017}.

Table~\ref{tab:dataset} presents the number or normal and adversarial samples used for training and testing the detection model (for both the CIFAR-10 and MNIST datasets). 

 \begin{table}[h]
 \centering
 \begin{tabular}{lllr}
 \toprule
  Dataset & Train/Test & \# Normal & \# Adversarial \\
 \midrule
  cifar10 &      train &        19463 &  10134 \\
  cifar10 &       test &        9339 &  7995 \\
   mnist &      train &        27239 &  26500 \\
   mnist &       test &        9910 &   9679 \\
 \bottomrule
 \end{tabular}
 \caption{Datasets description.}
 \label{tab:dataset}
 \end{table}

We used a variety of attack methods (see Figure~\ref{fig:RQ1-RQ2A}) for generating the adversarial examples. 

\subsubsection{SHAP values computation.}
For the [CIFAR,ResNet] model, we compute SHAP values on the last (and only) fully connected layer of the model which has a size of 64 neurons. 
This gives a total of 640 features per sample (64 features for each one of the ten target classes).
For the [MNIST,CNN] model, we compute SHAP values on the last fully connected layer of the model, which has a size of 128 neurons, and a total of 1280 features per sample.

\subsubsection{Training the adversarial example detector.}
In each experiment we generated an experiment-specific train and test sets, by fetching the relevant SHAP values from our generated repository. 
Using the SHAP values as features, we trained a fully connected feed forward neural network with three hidden layers (having 256, 128, 16 neurons respectively), all with ReLU activation units and Sigmoid output. 
We split the training set into train and validation subsets using a random 80/20 split and used the AdaBound optimizer~\cite{Luo2019AdaBound} with default parameters. 
We train for at most 500 epochs with an early stop condition that monitors the binary cross-entropy validation loss and decides to stop if it hasn't improved for the last 20 epochs.

\subsection{Results and Discussion}
\subsubsection{RQ1: baseline performance of the detector.}
\begin{figure}[h]
    \centering
    \includegraphics[scale=0.62]{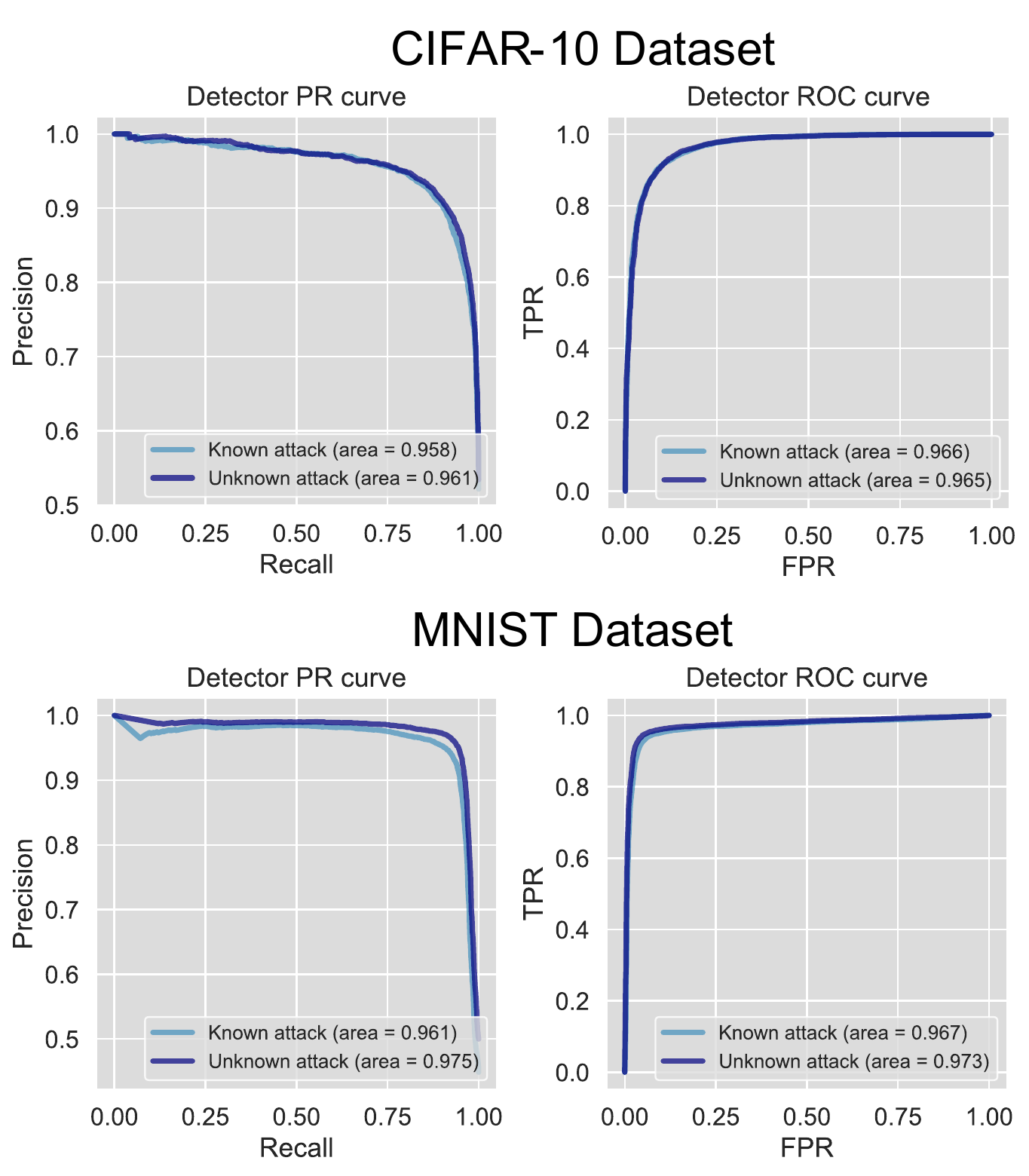}
    \caption{ROC and PR curves of the proposed detector evaluated on CIFAR-10 and MNIST datasets.}
    \label{fig:ROC-PR}
\end{figure}

For each dataset and model pair (i.e., [CIFAR,ResNet56] and [MNIST,CNN]) we constructed the train and test set as follows.
For the train set we used all normal and adversarial train samples of the selected dataset -- CIFAR-10 or MNIST.
Similarly, for the test set we used all normal and adversarial test samples of the selected dataset. 
Each sample in the train/test sets was represented using its SHAP values with the class label set to be ``1" for adversarial example and ``0" otherwise.
In Figure~\ref{fig:ROC-PR}, we present the ROC and precision-recall curves as well as the area under those curves (AUC-ROC and AUC-PR). 
As can be seen, the proposed method yields high detection performances with AUC-ROC of 0.966/0.967 and AUC-PR of 0.958/0.961 for the CIFAR-10 and MNIST datasets respectively.

\begin{figure}[h]
    \centering
    \includegraphics[scale=0.63]{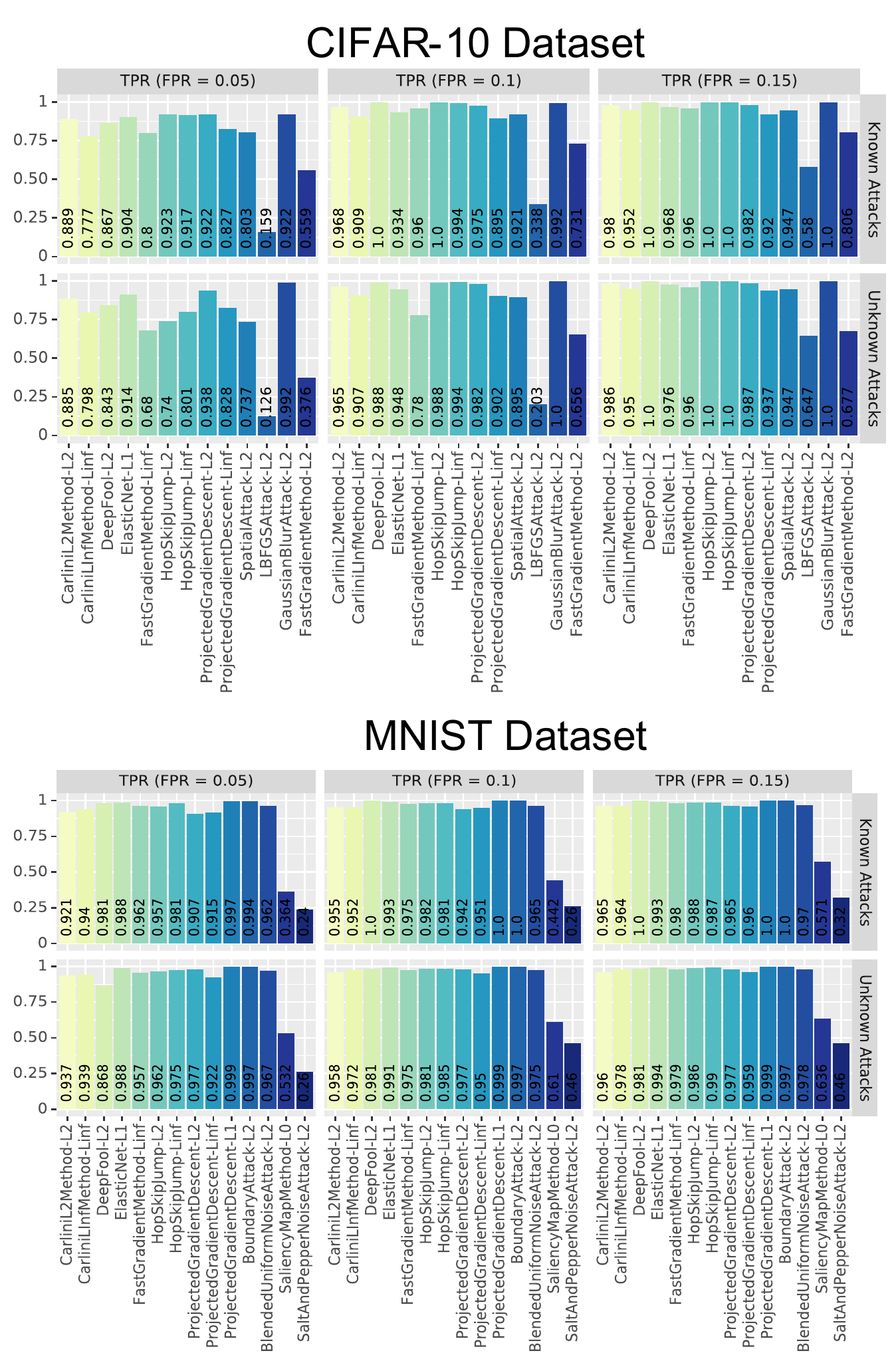}
    \caption{Evaluation results for RQ1 and RQ2 experiments.}
    \label{fig:RQ1-RQ2A}
\end{figure}

We also explore the specific detection rates of different attack methods.
In Figure~\ref{fig:RQ1-RQ2A}, we present the TPR of the detector for each attack method. The results show a high TPR for most attacks even for a FPR of 0.05.

\subsubsection{RQ2: Generalization across different attack types.}
This evaluation simulates a scenario where a detector trained on adversarial examples of known attacks is confronted with adversarial examples generated by an unknown attack.

Given a target dataset and model, we divided each one of its train and test sets by an (algorithm, metric) pair to get a collection of train/test subsets. 
We performed random under-sampling of each test subset to balance the number of normal and adversarial examples in it.
Then, we follow a ``leave on out" approach in which for each (algorithm,metric) pair we train a detector on all train samples generated by all but this pair and evaluate only on adversarial examples generated by this pair. 
Similar to the previous research question, we evaluate the general performance of our detector (Figure~\ref{fig:ROC-PR}), and the performances for each attack method separately (Figure~\ref{fig:ROC-PR}).
As can be seen the overall performance in detecting unknown attack is very similar to the case of known attacks (with AUC-ROC of 0.965/0.973 and AUC-PR of 0.961/0.975 for the CIFAR-10 and MNIST datasets respectively). 
Similarly, the proposed method show high performances for most attacks even for a FPR of 0.05.

\section{\label{sec:conclusions}Conclusions and Future Work}
The results of our experiments validate the ability of our approach to detect adversarial examples generated by a variety of state of the art attacks (RQ1).
We showed that the detector generalizes well when confronted with adversarial examples generated by attacks it wasn't train on (RQ2).
These results support our hypothesis regarding the connection between patterns of SHAP values of the penultimate layer of the classification model, the distribution of the importance of robust vs. non-robust features for the classification results and the ability to detect adversarial examples.

Although our detection method is based on a supervised learning model, which requires generating a big training set of adversarial examples using various attacks, the good generalization results imply that it should be possible to devise a semi-supervised detection approach based on the same features to streamline the detector training process and improve its performance further.

Our proposed method can be further extended into a generic framework, reminiscent of antivirus or IDS systems that continuously collect and analyze benign and malicious samples, extract signatures and, based on those signatures, classify samples as malicious or benign or forward them for manual analysis.
A framework such as this, employing both the SHAP based signatures discussed in this paper, and signatures used in other, state of the art detectors, could advance the practical ability to defend against adversarial examples.

In this paper we made a first step towards understanding the connection between model explanations and feature robustness. 
Rigorously studying this connection could be beneficial both for improving the performance of our detector and for a better understanding of the nature of adversarial examples.

Additional future work may include: (1) testing of our method on additional datasets (from other domains) and classification models; (2) evaluating the transferability of the detector between underlying classification models; and (3) evaluating our method against customized attacks adapted to take our detector into account (which is not trivial since the computation of SHAP values is not differentiable).

\bibliographystyle{IEEEtran}
\bibliography{references}

\begin{thebibliography}{10}
\providecommand{\url}[1]{#1}
\csname url@samestyle\endcsname
\providecommand{\newblock}{\relax}
\providecommand{\bibinfo}[2]{#2}
\providecommand{\BIBentrySTDinterwordspacing}{\spaceskip=0pt\relax}
\providecommand{\BIBentryALTinterwordstretchfactor}{4}
\providecommand{\BIBentryALTinterwordspacing}{\spaceskip=\fontdimen2\font plus
\BIBentryALTinterwordstretchfactor\fontdimen3\font minus
  \fontdimen4\font\relax}
\providecommand{\BIBforeignlanguage}[2]{{%
\expandafter\ifx\csname l@#1\endcsname\relax
\typeout{** WARNING: IEEEtran.bst: No hyphenation pattern has been}%
\typeout{** loaded for the language `#1'. Using the pattern for}%
\typeout{** the default language instead.}%
\else
\language=\csname l@#1\endcsname
\fi
#2}}
\providecommand{\BIBdecl}{\relax}
\BIBdecl

\bibitem{szegedy2013intriguing}
C.~Szegedy, W.~Zaremba, I.~Sutskever, J.~Bruna, D.~Erhan, I.~Goodfellow, and
  R.~Fergus, ``Intriguing properties of neural networks,'' \emph{arXiv preprint
  arXiv:1312.6199}, 2013.

\bibitem{moosavi2016deepfool}
S.-M. Moosavi-Dezfooli, A.~Fawzi, and P.~Frossard, ``Deepfool: a simple and
  accurate method to fool deep neural networks,'' in \emph{Proceedings of the
  IEEE conference on computer vision and pattern recognition}, 2016, pp.
  2574--2582.

\bibitem{athalye2017synthesizing}
A.~Athalye, L.~Engstrom, A.~Ilyas, and K.~Kwok, ``Synthesizing robust
  adversarial examples,'' \emph{arXiv preprint arXiv:1707.07397}, 2017.

\bibitem{carlini2017towards}
N.~Carlini and D.~Wagner, ``Towards evaluating the robustness of neural
  networks,'' in \emph{2017 IEEE Symposium on Security and Privacy (SP)}.\hskip
  1em plus 0.5em minus 0.4em\relax IEEE, 2017, pp. 39--57.

\bibitem{kurakin2016adversarial}
A.~Kurakin, I.~Goodfellow, and S.~Bengio, ``Adversarial machine learning at
  scale,'' 2016.

\bibitem{madry2017towards}
A.~Madry, A.~Makelov, L.~Schmidt, D.~Tsipras, and A.~Vladu, ``Towards deep
  learning models resistant to adversarial attacks,'' \emph{arXiv preprint
  arXiv:1706.06083}, 2017.

\bibitem{papernot2016distillation}
N.~Papernot, P.~McDaniel, X.~Wu, S.~Jha, and A.~Swami, ``Distillation as a
  defense to adversarial perturbations against deep neural networks,'' in
  \emph{2016 IEEE Symposium on Security and Privacy (SP)}.\hskip 1em plus 0.5em
  minus 0.4em\relax IEEE, 2016, pp. 582--597.

\bibitem{dziugaite2016study}
G.~K. Dziugaite, Z.~Ghahramani, and D.~M. Roy, ``A study of the effect of jpg
  compression on adversarial images,'' \emph{arXiv preprint arXiv:1608.00853},
  2016.

\bibitem{shaham2018understanding}
U.~Shaham, Y.~Yamada, and S.~Negahban, ``Understanding adversarial training:
  Increasing local stability of supervised models through robust
  optimization,'' \emph{Neurocomputing}, vol. 307, pp. 195--204, 2018.

\bibitem{metzen2017detecting}
J.~H. Metzen, T.~Genewein, V.~Fischer, and B.~Bischoff, ``On detecting
  adversarial perturbations,'' \emph{arXiv preprint arXiv:1702.04267}, 2017.

\bibitem{feinman2017detecting}
R.~Feinman, R.~R. Curtin, S.~Shintre, and A.~B. Gardner, ``Detecting
  adversarial samples from artifacts,'' \emph{arXiv preprint arXiv:1703.00410},
  2017.

\bibitem{grosse2017statistical}
K.~Grosse, P.~Manoharan, N.~Papernot, M.~Backes, and P.~McDaniel, ``On the
  (statistical) detection of adversarial examples,'' \emph{arXiv preprint
  arXiv:1702.06280}, 2017.

\bibitem{xu2017feature}
W.~Xu, D.~Evans, and Y.~Qi, ``Feature squeezing: Detecting adversarial examples
  in deep neural networks,'' \emph{arXiv preprint arXiv:1704.01155}, 2017.

\bibitem{katzir2018detecting}
Z.~Katzir and Y.~Elovici, ``Detecting adversarial perturbations through spatial
  behavior in activation spaces,'' \emph{arXiv preprint arXiv:1811.09043},
  2018.

\bibitem{pang2018towards}
T.~Pang, C.~Du, Y.~Dong, and J.~Zhu, ``Towards robust detection of adversarial
  examples,'' in \emph{Advances in Neural Information Processing Systems},
  2018, pp. 4579--4589.

\bibitem{roth2019odds}
K.~Roth, Y.~Kilcher, and T.~Hofmann, ``The odds are odd: A statistical test for
  detecting adversarial examples,'' in \emph{International Conference on
  Machine Learning}, 2019, pp. 5498--5507.

\bibitem{shamir2019simple}
A.~Shamir, I.~Safran, E.~Ronen, and O.~Dunkelman, ``A simple explanation for
  the existence of adversarial examples with small hamming distance,''
  \emph{arXiv preprint arXiv:1901.10861}, 2019.

\bibitem{ilyas2019adversarial}
A.~Ilyas, S.~Santurkar, D.~Tsipras, L.~Engstrom, B.~Tran, and A.~Madry,
  ``Adversarial examples are not bugs, they are features,'' \emph{arXiv
  preprint arXiv:1905.02175}, 2019.

\bibitem{goodfellow2014explaining}
I.~J. Goodfellow, J.~Shlens, and C.~Szegedy, ``Explaining and harnessing
  adversarial examples,'' \emph{arXiv preprint arXiv:1412.6572}, 2014.

\bibitem{carlini2017adversarial}
N.~Carlini and D.~Wagner, ``Adversarial examples are not easily detected:
  Bypassing ten detection methods,'' in \emph{Proceedings of the 10th ACM
  Workshop on Artificial Intelligence and Security}.\hskip 1em plus 0.5em minus
  0.4em\relax ACM, 2017, pp. 3--14.

\bibitem{gunning2017explainable}
D.~Gunning, ``Explainable artificial intelligence (xai),'' \emph{Defense
  Advanced Research Projects Agency (DARPA), nd Web}, vol.~2, 2017.

\bibitem{adadi2018peeking}
A.~Adadi and M.~Berrada, ``Peeking inside the black-box: A survey on
  explainable artificial intelligence (xai),'' \emph{IEEE Access}, vol.~6, pp.
  52\,138--52\,160, 2018.

\bibitem{samek2017explainable}
W.~Samek, T.~Wiegand, and K.-R. M{\"u}ller, ``Explainable artificial
  intelligence: Understanding, visualizing and interpreting deep learning
  models,'' \emph{arXiv preprint arXiv:1708.08296}, 2017.

\bibitem{zhang2018visual}
Q.-s. Zhang and S.-C. Zhu, ``Visual interpretability for deep learning: a
  survey,'' \emph{Frontiers of Information Technology \& Electronic
  Engineering}, vol.~19, no.~1, pp. 27--39, 2018.

\bibitem{shap2017}
\BIBentryALTinterwordspacing
S.~M. Lundberg and S.-I. Lee, ``A unified approach to interpreting model
  predictions,'' in \emph{Advances in Neural Information Processing Systems
  30}, I.~Guyon, U.~V. Luxburg, S.~Bengio, H.~Wallach, R.~Fergus,
  S.~Vishwanathan, and R.~Garnett, Eds.\hskip 1em plus 0.5em minus 0.4em\relax
  Curran Associates, Inc., 2017, pp. 4765--4774. [Online]. Available:
  \url{http://papers.nips.cc/paper/7062-a-unified-approach-to-interpreting-model-predictions.pdf}
\BIBentrySTDinterwordspacing

\bibitem{krizhevsky2009learning}
A.~Krizhevsky, G.~Hinton \emph{et~al.}, ``Learning multiple layers of features
  from tiny images,'' Citeseer, Tech. Rep., 2009.

\bibitem{lecun1998gradient}
Y.~LeCun, L.~Bottou, Y.~Bengio, P.~Haffner \emph{et~al.}, ``Gradient-based
  learning applied to document recognition,'' \emph{Proceedings of the IEEE},
  vol.~86, no.~11, pp. 2278--2324, 1998.

\bibitem{carlini2019evaluating}
N.~Carlini, A.~Athalye, N.~Papernot, W.~Brendel, J.~Rauber, D.~Tsipras,
  I.~Goodfellow, and A.~Madry, ``On evaluating adversarial robustness,''
  \emph{arXiv preprint arXiv:1902.06705}, 2019.

\bibitem{papernot2017practical}
N.~Papernot, P.~McDaniel, I.~Goodfellow, S.~Jha, Z.~B. Celik, and A.~Swami,
  ``Practical black-box attacks against machine learning,'' in
  \emph{Proceedings of the 2017 ACM on Asia conference on computer and
  communications security}.\hskip 1em plus 0.5em minus 0.4em\relax ACM, 2017,
  pp. 506--519.

\bibitem{ribeiro2016should}
\BIBentryALTinterwordspacing
M.~T. Ribeiro, S.~Singh, and C.~Guestrin, ``“why should i trust you?”,''
  \emph{Proceedings of the 22nd ACM SIGKDD International Conference on
  Knowledge Discovery and Data Mining - KDD ’16}, 2016. [Online]. Available:
  \url{http://dx.doi.org/10.1145/2939672.2939778}
\BIBentrySTDinterwordspacing

\bibitem{shrikumar2017learning}
A.~Shrikumar, P.~Greenside, and A.~Kundaje, ``Learning important features
  through propagating activation differences,'' 2017.

\bibitem{goodfellow2016dlbook}
I.~Goodfellow, Y.~Bengio, and A.~Courville, \emph{Deep Learning}.\hskip 1em
  plus 0.5em minus 0.4em\relax MIT Press, 2016,
  \url{http://www.deeplearningbook.org}.

\bibitem{mcinnes2018arXivUMAP}
L.~{McInnes}, J.~{Healy}, and J.~{Melville}, ``{UMAP: Uniform Manifold
  Approximation and Projection for Dimension Reduction},'' \emph{ArXiv
  e-prints}, Feb. 2018.

\bibitem{he2016deep}
K.~He, X.~Zhang, S.~Ren, and J.~Sun, ``Deep residual learning for image
  recognition,'' in \emph{Proceedings of the IEEE conference on computer vision
  and pattern recognition}, 2016, pp. 770--778.

\bibitem{rauber2017foolbox}
\BIBentryALTinterwordspacing
J.~Rauber, W.~Brendel, and M.~Bethge, ``Foolbox: A python toolbox to benchmark
  the robustness of machine learning models,'' \emph{arXiv preprint
  arXiv:1707.04131}, 2017. [Online]. Available:
  \url{http://arxiv.org/abs/1707.04131}
\BIBentrySTDinterwordspacing

\bibitem{art2018}
\BIBentryALTinterwordspacing
M.-I. Nicolae, M.~Sinn, M.~N. Tran, B.~Buesser, A.~Rawat, M.~Wistuba,
  V.~Zantedeschi, N.~Baracaldo, B.~Chen, H.~Ludwig, I.~Molloy, and B.~Edwards,
  ``Adversarial robustness toolbox v0.10.0,'' \emph{CoRR}, vol. 1807.01069,
  2018. [Online]. Available: \url{https://arxiv.org/pdf/1807.01069}
\BIBentrySTDinterwordspacing

\bibitem{Luo2019AdaBound}
L.~Luo, Y.~Xiong, Y.~Liu, and X.~Sun, ``Adaptive gradient methods with dynamic
  bound of learning rate,'' in \emph{Proceedings of the 7th International
  Conference on Learning Representations}, New Orleans, Louisiana, May 2019.

\end{thebibliography}

\end{document}